\def\year{2020}\relax
\documentclass[letterpaper]{article} 
\usepackage{aaai20}  
\usepackage{times}  
\usepackage{helvet} 
\usepackage{courier}  
\usepackage[hyphens]{url}  
\usepackage{graphicx} 
\usepackage{graphicx}  

\usepackage{amsmath}
\usepackage{amssymb}
\usepackage{subfigure}
\usepackage{epsfig}
\usepackage{epstopdf}
\usepackage{algorithm}
\usepackage{algorithmic}
\usepackage{multirow}

\title{Learning Attentive Pairwise Interaction for Fine-Grained Classification}
\author{Peiqin Zhuang,\thanks{Equally-contributed first authors}\textsuperscript{\rm 1,2} Yali Wang,\textsuperscript{$*$\rm 1,2} Yu Qiao\thanks{Corresponding author}\textsuperscript{\rm 1,2}\\
\textsuperscript{\rm 1}ShenZhen Key Lab of Computer Vision and Pattern Recognition,
SIAT-SenseTime Joint Lab, \\
Shenzhen Institutes of Advanced Technology, Chinese Academy of Sciences \\
\textsuperscript{\rm 2}SIAT Branch, Shenzhen Institute of Artificial Intelligence and Robotics for Society\\
\{pq.zhuang, yl.wang, yu.qiao\}@siat.ac.cn
}

\begin{document}

\maketitle

\begin{abstract}
Fine-grained classification is a challenging problem,
due to subtle differences among highly-confused categories.
Most approaches address this difficulty by learning discriminative representation of individual input image.
On the other hand,
humans can effectively identify contrastive clues by comparing image pairs.
Inspired by this fact,
this paper proposes a simple but effective Attentive Pairwise Interaction Network (API-Net),
which can progressively recognize a pair of fine-grained images by interaction.
Specifically,
API-Net first learns a mutual feature vector to capture semantic differences in the input pair.
It then compares this mutual vector with individual vectors to generate gates for each input image.
These distinct gate vectors inherit mutual context on semantic differences,
which allow API-Net to attentively capture contrastive clues by pairwise interaction between two images.
Additionally,
we train API-Net in an end-to-end manner with a score ranking regularization,
which can further generalize API-Net by taking feature priorities into account.
We conduct extensive experiments on five popular benchmarks in fine-grained classification.
API-Net outperforms the recent SOTA methods,
i.e.,
CUB-200-2011 (90.0\%),
Aircraft (93.9\%),
Stanford Cars (95.3\%),
Stanford Dogs (90.3\%),
and NABirds (88.1\%).
\end{abstract}

\begin{figure}[htb!]
\centering

\includegraphics[width=0.9\columnwidth]{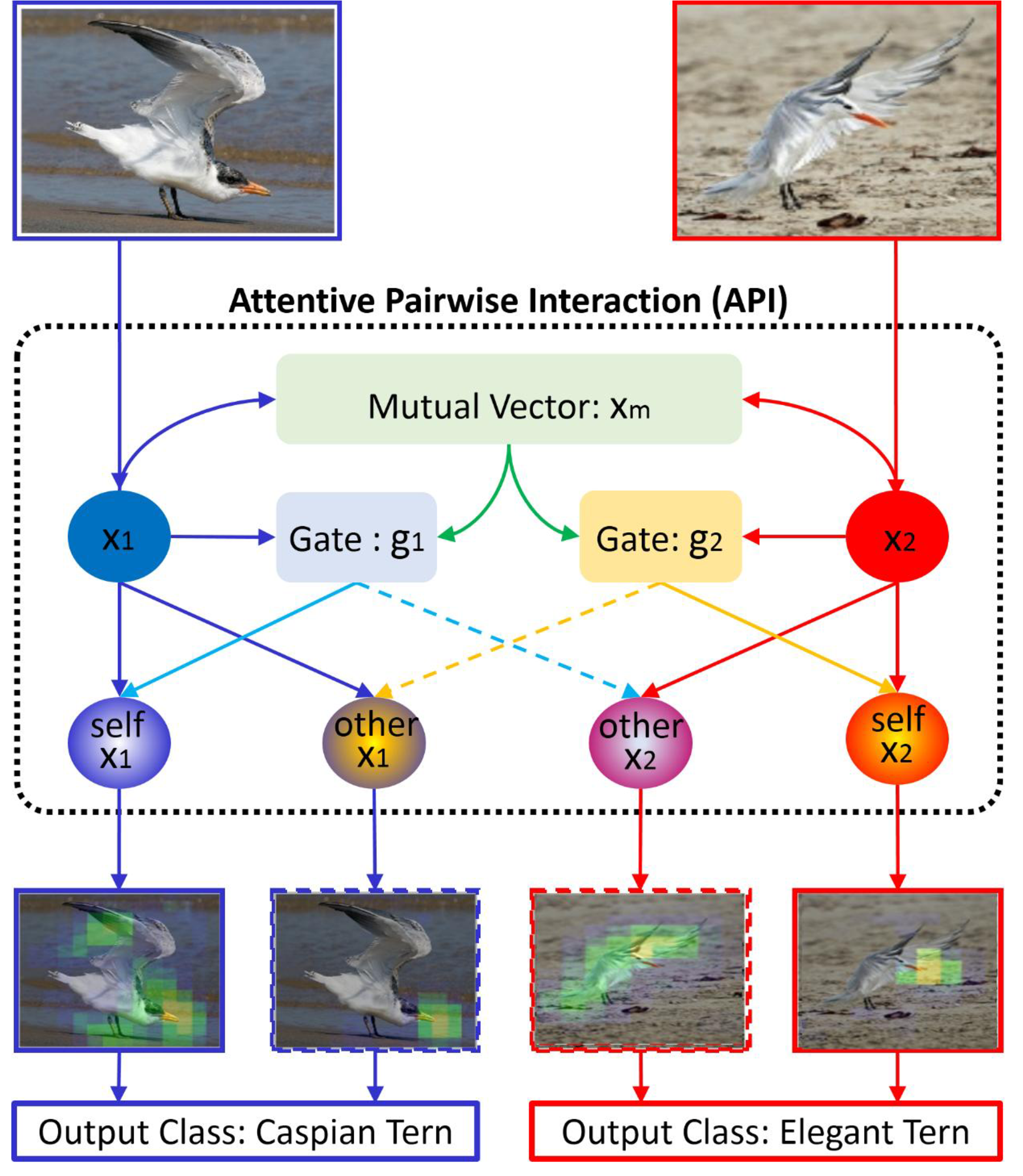}
\caption{
Motivation (Best view in color).
\textit{Caspian Tern} and \textit{Elegant Tern} are two highly-confused bird species.
Humans often distinguish them by pairwise comparison,
instead of checking each individual image alone.
First,
humans would exploit contrastive clues (e.g. body and mouth) from the image pair,
and then check each image with these mutual contexts to further discover distinct attentions on each image.
Finally,
humans recognize both images via comparing subtle differences jointly.
To mimic this capacity,
we propose a novel API-Net.
More details can be found in Section \ref{Introduction}.}
\label{fig:show}
\end{figure}

\section{Introduction}
\label{Introduction}

Over the past years,
CNNs have achieved remarkable successes for visual recognition \cite{he2016deep,huang2017densely}.
However,
these classical models are often limited to distinguish fine-grained categories,
due to highly-confused appearances.
Subsequently,
a number of fine-grained frameworks have been proposed by
finding key part regions \cite{fu2017look,zheng2017learning,yang2018learning},
learning patch relations \cite{lin2015bilinear,cai2017higher,yu2018hierarchical},
etc.
But most of them take individual image as input,
which may limit their ability to identify contrastive clues from different images for fine-grained classification.

On the contrary,
humans often recognize fine-grained objects by comparing image pairs \cite{bruner2017study},
instead of checking single images alone.
For example,
\textit{Caspian Tern} and \textit{Elegant Tern} are two highly-confused bird species,
as shown in Fig. \ref{fig:show}.
If we only check individual image alone,
it is difficult to recognize which categories it belongs to,
especially when the bird is self-occluded with a noisy background.
Alternatively,
we often take a pair of images together,
and summarize contrastive visual appearances as context,
e.g.,
body, mouth and etc.
Then,
we check individual images with these mutual contexts,
so that we can further understand distinct aspects of each image,
e.g.,

the body of the bird is an important part for the left image,
while the mouth of the bird is a key characteristic for the right image.
With such discriminative guidance,
we pay different attentions to the body and the mouth of two birds.
Note that,
for each bird in the pair,
we not only check its prominent part but also have a glance at the distinct part found from the other bird.
This comparative interaction can effectively tell us that,
\textit{Caspian Tern} has a fatter body,
while
\textit{Elegant Tern} has a more curved mouth.
Consequently,
we recognize both images jointly.

To mimic this capacity of human beings,
we introduce a novel Attentive Pairwise Interaction Network (API-Net) for fine-grained classification.
It can adaptively discover contrastive clues from a pair of fine-grained images,
and attentively distinguish them via pair interaction.

More specifically,
API can effectively recognize two fine-grained images,
by a progressive comparison procedure like human beings.
To achieve this goal,
API-Net consists of three submodules,
i.e.,
mutual vector learning,
gate vector generation,
and
pairwise interaction.
By taking a pair of fine-grained images as input,

API-Net first learns a mutual vector to summarize contrastive clues of input pair as context.
Then,
it compares mutual vector with individual vectors.
This allows API-Net to generate distinct gates,
which can effectively highlight semantic differences respectively from the view of each individual image.

Consequently,
API-Net uses these gates as discriminative attentions to perform pairwise interaction.
In this case,
each image can generate two enhanced feature vectors,
which are activated respectively from its own gate vector and the gate vector of the other image in the pair.
Via an end-to-end training manner with a score-ranking regularization,
API-Net can promote the discriminative ability of all these features jointly with different priorities.
Additionally,
it is worth mentioning that,
one can easily embed API into any CNN backbones for fine-grained classification,
and flexibly unload it for single-input test images without loss of generalization capacity.
Such plug-and-play property makes API as a preferable choice in practice.
Finally,
we evaluate API-Net on five popular benchmarks in fine-grained recognition,
namely CUB-200-2011, Aircraft, Stanford Cars, Stanford Dogs and NABirds.
The extensive results show that,
API-Net achieves the state-of-the-art performance on all these datasets.

\section{Related Works}

A number of research works have been recently proposed for fine-grained classification.
In the following,
we mainly summarize and discuss those related works.

\textbf{Object Parts Localization}.
These approaches mainly utilize the pre-defined bounding boxes or part annotations to capture visual details in the local regions 
\cite{zhang2014part,lin2015deep}.
However,
collecting such annotations is often labor-intensive or infeasible in practice.
Hence,
several weakly-supervised localization approaches have been recently proposed by
designing complex spatial attention mechanisms (e.g., RA-CNN \cite{fu2017look}, MA-CNN \cite{zheng2017learning}),
learning a bank of discriminative filters \cite{wang2018learning},
guiding region detection with multi-agent cooperation \cite{yang2018learning},
etc.
But these approaches mainly focus on mining local characteristics of fine-grained images.
Consequently,
they may lack the capacity of discriminative feature learning.


\textbf{Discriminative Feature Learning}.
To learn the representative features,
many approaches have been exploited by patch interactions \cite{lin2015bilinear,cai2017higher,yu2018hierarchical}.
A well-known approach is B-CNN \cite{lin2015bilinear},
which performs bilinear pooling on the representations of two local patches in an image.
Following this direction,
several high-order approaches have been proposed via
polynomial kernel formulation \cite{cai2017higher},
cross-layer bilinear representation \cite{yu2018hierarchical},
etc.
However,
these approaches take a single image alone as input,
while neglecting comparisons between different images,
i.e.,
an important clue to distinguish highly-confused objects.

\textbf{Metric Learning}.
Metric learning refers to the method that uses similarity measurements to model relations between image pairs \cite{kulis2013metric}.
It has been widely used in
Face Verification \cite{schroff2015facenet},
Person ReID \cite{hermans2017defense} and so on.
Recently,
it has been introduced for fine-grained classification,
e.g.,
triplet loss design \cite{zhang2016embedding},
pairwise confusion regularization \cite{dubey2018pairwise},
multi-attention multi-class constraint \cite{sun2018multi},
etc.
However,
these approaches mainly leverage metric learning to improve sample distributions in the feature space.
Hence,
they often lack the adaptation capacity,
w.r.t.,
how to discover visual differences between a pair of images.

Different from previous approaches,
our API-Net can adaptively summarize contrastive clues,
by learning a mutual vector from an image pair. 
As a result,
we can leverage it as guidance,
and attentively distinguish two fine-grained images via pairwise interaction.

\begin{figure*}[t]
\centering
\includegraphics[width=1.96\columnwidth]{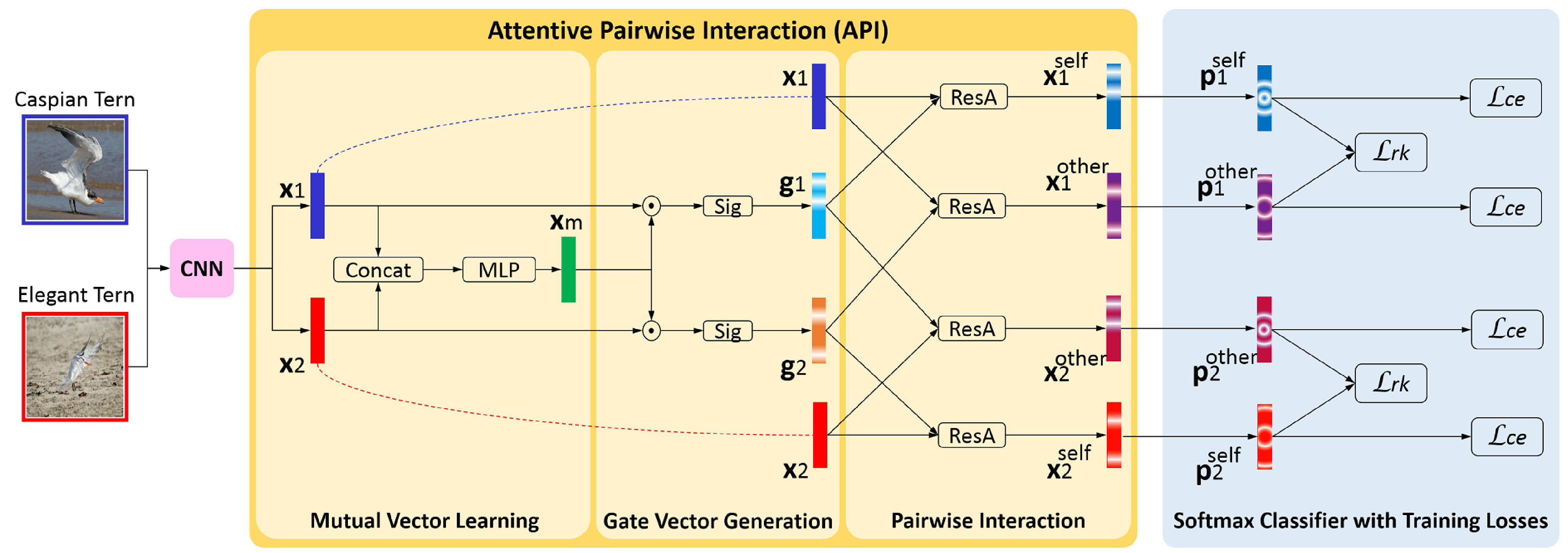}
\caption{The framework of API-Net (Best view in color).
API can progressively recognize a pair of fine-grained images,
based on a novel human-like learning procedure.
It consists of three submodules.
\textbf{1. Mutual Vector Learning}.
API learns a mutual vector $\mathbf{x}_{m}$ from individual $\mathbf{x}_{1}$ and $\mathbf{x}_{2}$ (Eq. \ref{eq:MutualVector}).
In this case,
it can summarize contrastive cues in the pair.
\textbf{2. Gate Vector Generation}.
API further compares $\mathbf{x}_{m}$ with $\mathbf{x}_{1}$ and $\mathbf{x}_{2}$ ,
and generates two distinct gate vectors $\mathbf{g}_{i}$ (Eq. \ref{eq:GateVector}).
These gates allow API to discover distinct clues respectively from the view of each individual image.
\textbf{3. Pairwise Interaction}.
API performs pairwise interaction with guidance of gate vectors (Eq. \ref{eq:InterVector11}-\ref{eq:InterVector21}).
Via training API-Net with a score-ranking regularization,
we can distinguish all these features jointly with the consideration of feature priorities (Eq. \ref{eq:LossTotal}-\ref{eq:LossRK}).
Additionally,
API is a practical plug-and-play module,
i.e.,
one can combine API with CNNs during training, and flexibly unload it for single-input test images.
}
\label{fig:API-Net}
\end{figure*}

\section{Attentive Pairwise Interaction}
\label{Attentive Pair Interaction Network}

In this section,
we describe Attentive Pairwise Interaction Network (API-Net) for fine-grained classification.
Our design is partially inspired by the observation that,
instead of learning visual concepts alone with individual image,
humans often compare a pair of images jointly to distinguish subtle differences between similar objects \cite{bruner2017study}.

To imitate this capacity,
we simultaneously take a pair of images as input to API-Net,
and progressively distinguish them by three elaborative submodules,
i.e.,
mutual vector learning,
gate vector generation,
and pairwise interaction.
The whole framework is demonstrated in Fig. \ref{fig:API-Net}.

\textbf{Mutual Vector Learning}.
First,
we feed two fine-grained images into a CNN backbone,
and extract their $D$-dimension feature vectors respectively,
i.e.,
$\mathbf{x}_{1}$ and $\mathbf{x}_{2}\in\mathbb{R}^{D}$.
Then,
we learn a mutual vector $\mathbf{x}_{m}\in\mathbb{R}^{D}$ from individual $\mathbf{x}_{1}$ and $\mathbf{x}_{2}$,
\begin{equation}
\mathbf{x}_{m}=f_{m}([\mathbf{x}_{1},\mathbf{x}_{2}]),
\label{eq:MutualVector}
\end{equation}
where
$f_{m}(\cdot)$ is a mapping function of $[\mathbf{x}_{1},\mathbf{x}_{2}]$,
e.g.,
a simple MLP works well in our experiments.
Since $\mathbf{x}_{m}$ is adaptively summarized from both $\mathbf{x}_{1}$ and $\mathbf{x}_{2}$,
it often contains feature channels which indicate high-level contrastive clues (e.g. body and mouth of two birds in Fig.\ref{fig:show}) in the pair.

\textbf{Gate Vector Generation}.
After learning the mutual vector $\mathbf{x}_{m}$,
we propose to compare it with $\mathbf{x}_{1}$ and $\mathbf{x}_{2}$.
The main reason is that,
we should further generate distinct clues respectively from the view of each individual image,
in order to distinguish this pair afterwards.

In particular,
we perform channel-wise product between $\mathbf{x}_{m}$ and $\mathbf{x}_{i}$,
so that we can leverage $\mathbf{x}_{m}$ as guidance to find which channels of individual $\mathbf{x}_{i}$ may contain contrastive clues.
Then,
we add a sigmoid function to generate the gate vector $\mathbf{g}_{i}\in\mathbb{R}^{D}$,
\begin{equation}
\mathbf{g}_{i}=sigmoid(\mathbf{x}_{m}\odot\mathbf{x}_{i}),~~i\in\{1,2\}.
\label{eq:GateVector}
\end{equation}
As a result,
$\mathbf{g}_{i}$ becomes a discriminative attention which highlights semantic differences with a distinct view of each individual $\mathbf{x}_{i}$,
e.g.,
the body of the bird is important in one image of Fig.\ref{fig:show},
while the mouth of the bird is a key in the other image.
In our experiments,
we evaluate different gating strategies and show effectiveness of our design.

\textbf{Pairwise Interaction}.
Next,
we perform pairwise interaction by gate vectors.
Our design is partially motivated by the fact that,
to capture subtle differences in a pair of fine-grained images,
human check each image not only with its prominent parts but also with distinct parts from the other image.
For this reason,
we introduce an interaction mechanism via residual attention,
\begin{align}
\mathbf{x}_{1}^{self}&=\mathbf{x}_{1}+\mathbf{x}_{1}\odot\mathbf{g}_{1},\label{eq:InterVector11}\\
\mathbf{x}_{2}^{self}&=\mathbf{x}_{2}+\mathbf{x}_{2}\odot\mathbf{g}_{2},\label{eq:InterVector22}\\
\mathbf{x}_{1}^{other}&=\mathbf{x}_{1}+\mathbf{x}_{1}\odot\mathbf{g}_{2},\label{eq:InterVector12}\\
\mathbf{x}_{2}^{other}&=\mathbf{x}_{2}+\mathbf{x}_{2}\odot\mathbf{g}_{1}.\label{eq:InterVector21}
\end{align}
As we can see,
each individual feature $\mathbf{x}_{i}$ in the pair produces two attentive feature vectors,
i.e.,
$\mathbf{x}_{i}^{self}\in\mathbb{R}^{D}$ is highlighted by its own gate vector,
and
$\mathbf{x}_{i}^{other}\in\mathbb{R}^{D}$ is activated by the gate vector of the other image in the pair.
In this case,
we enhance $\mathbf{x}_{i}$ with discriminative clues that come from both images.
Via distinguishing all these features jointly,
we can reduce confusion in this fine-grained pair.

\textbf{Training}.
After obtaining four attentive features $\mathbf{x}_{i}^{j}$ in the pair (where $i\in\{1,2\}$, $j\in\{self,other\}$),
we feed them into a $softmax$ classifier,
\begin{equation}
\mathbf{p}_{i}^{j}=softmax(\mathbf{W}\mathbf{x}_{i}^{j}+\mathbf{b}), 
\label{eq:prob}
\end{equation}
where
$\mathbf{p}_{i}^{j}\in\mathbb{R}^{C}$ is the prediction score vector,
$C$ is the number of object categories,
and
$\{\mathbf{W}, \mathbf{b}\}$ is the parameter set of classifier.
To train the whole API-Net effectively,
we design the following loss $\mathcal{L}$ for a pair,
\begin{equation}
\mathcal{L}=\mathcal{L}_{ce}+\lambda \mathcal{L}_{rk},
\label{eq:LossTotal}
\end{equation}
where
$\mathcal{L}_{ce}$ is a cross entropy loss,
and $\mathcal{L}_{rk}$ is a score ranking regularization with coefficient $\lambda$.

\textbf{1) Cross Entropy Loss}.
The main loss is the cross entropy loss $\mathcal{L}_{ce}$,
\begin{equation}
\mathcal{L}_{ce}=-\sum\nolimits_{i\in\{1,2\}}\sum\nolimits_{j\in\{self,other\}}\mathbf{y}_{i}^{\top}\log(\mathbf{p}_{i}^{j}),
\label{eq:LossCE}
\end{equation}
where
$\mathbf{y}_{i}$ is the one-hot label vector for image $i$ in the pair,
$\top$ denotes the vector transposition.
By using this loss,
API-Net can gradually recognize all the attentive features $\mathbf{x}_{i}^{j}$,
with supervision of label $\mathbf{y}_{i}$.

\textbf{2) Score Ranking Regularization}.
Additionally,
we introduce a hinge loss as the score ranking regularization $\mathcal{L}_{rk}$,
\begin{equation}
\mathcal{L}_{rk}=\sum\nolimits_{i\in\{1,2\}} \max(0,\mathbf{p}_{i}^{other}(c_{i})-\mathbf{p}_{i}^{self}(c_{i})+\epsilon),
\label{eq:LossRK}
\end{equation}
where
$\mathbf{p}_{i}^{j}(c_{i})\in \mathbb{R}$ is the score obtained from the prediction vector $\mathbf{p}_{i}^{j}$,
and $c_{i}$ is the corresponding index associated with the ground truth label of image $i$.
Our motivation of this design is that,
$x^{self}_{i}$ is activated by its own gate vector.
Hence,
it should be more discriminative to recognize the corresponding image,
compared with $x^{other}_{i}$.

To take this knowledge into account,
we use $\mathcal{L}_{rk}$ to promote the priority of $\mathbf{x}_{i}^{self}$,
i.e.,
the score difference $\mathbf{p}_{i}^{self}(c_{i})-\mathbf{p}_{i}^{other}(c_{i})$ should be larger than a margin $\epsilon$.
With such a regularization,
API-Net learns to recognize each image in the pair by adaptively taking feature priorities into account.

\textbf{3) Pair Construction}.
Eq. (\ref{eq:LossTotal}) defines the loss for a training pair.
Next,
we explain how to construct multiple pairs in a batch for end-to-end training.
Specifically,
we randomly sample $N_{cl}$ classes in a batch.
For each class,
we randomly sample $N_{im}$ training images.
Then,
we feed all these images into CNN backbone to generate their feature vectors.
For each image,
we compare its feature with others in the batch,
according to Euclidean distance.
As a result,
we can construct two pairs for each image,
i.e.,
the intra / inter pair consists of its feature and its most similar feature from intra / inter classes in the batch.
This design allows our API-Net for learning to distinguish which are highly-confused or truly-similar pairs.
We also investigate different construction strategies in our experiments.
Consequently,
there are $2\times N_{cl}\times N_{im}$ pairs in each batch.
We pass them into our API module,
and summarize the loss $\mathcal{L}$ over all these pairs for end-to-end training.

\textbf{Testing}.
We would like to emphasize that,
API is a practical plug-and-play module for fine-grained classification.
In the training phase,
this module can summarize contrastive clues from a pair,
which can gradually generalize the discriminative power of CNN representations for each individual image.
Hence,
in the testing phase,
one can simply unload API for single-input test images,
without much loss of generalization.
Specifically,
we feed a test image into the CNN backbone to extract its feature $\mathbf{x}_{\star}\in\mathbb{R}^{D}$.
Then,
we directly put $\mathbf{x}_{\star}$ into softmax classifier (Eq. \ref{eq:prob}).
The resulting score vector $\mathbf{p}_{\star}\in\mathbb{R}^{C}$ is used for label prediction.
By doing so,
our testing manner is just the same as that of a plain CNN,
which largely boosts the value of API-Net for fine-grained classification.

\section{Experiments}

\textbf{Data Sets}.
We evaluate API-Net on five popular fine-grained benchmarks,
i.e.,
CUB-200-2011\cite{wah2011caltech},
Aircraft\cite{maji2013fine},
Stanford Cars\cite{krause20133d},
Stanford Dogs\cite{khosla2011novel}
and
NABirds \cite{van2015building}.
Specifically,
CUB-200-2011 / Aircraft / Stanford Cars / Stanford Dogs / NABirds consists of 11,788 / 10,000 / 16,185 / 20,580 / 48,562 images,
from 200 bird / 100 airplane / 196 car / 120 dog / 555 bird classes.
We use the official train \& test splits for evaluation.


\textbf{Implementation Details}.
Unless stated otherwise,
we implement API-Net as follows.
First,
we resize each image as $512 \times 512$,
and crop a $448 \times 448$ patch as input to API-Net (train: random cropping, test: center cropping).
Furthermore,
we use ResNet-101 (pretrained on ImageNet) as CNN backbone,
and extract the feature vector $\mathbf{x}_{i}\in\mathbb{R}^{2048}$ after global pooling average operation.
Second,
for all the datasets,
we randomly sample 30 categories in each batch.
For each category,
we randomly sample 4 images.
For each image,
we find its most similar image from its own class and the rest classes,
according to Euclidean distance between features.
As a result,
we obtain an intra pair and an inter pair for each image in the batch.
For each pair,
we concatenate $\mathbf{x}_{1}$ and $\mathbf{x}_{2}$ as input to a two-layer MLP,
i.e., FC(4096$\rightarrow$512)-FC(512$\rightarrow$2048).
Consequently,
this operation generates the mutual vector $\mathbf{x}_{m}\in\mathbb{R}^{2048}$.
Finally,
we implement our network by PyTorch.
For all the datasets,
the coefficient $\lambda$ in Eq. (\ref{eq:LossTotal}) is 1.0,
and the margin $\epsilon$ in the score-ranking regularization is 0.05.
We use the standard SGD with
momentum of 0.9,
weight decay of 0.0005.
Furthermore,
the initial learning rate is 0.01 (0.001 for Stanford Dogs),
and adopt cosine annealing strategy to adjust it.
The total number of training epochs is 100 (50 for Stanford Dogs).
Besides, during training phase,
we freeze the conv layers and only train the newly-added fully-connected layers
in the first 8 epochs(12 epochs for Standord Dogs).

\subsection{Ablation Studies}
\label{Ablation Studies}

To investigate the properties of our API-Net,
we evaluate its key designs on CUB-200-2011.
For fairness,
when we explore different strategies of one design,
others are set as the basic strategy stated in the implementation details.

\textbf{Basic Methods}.
First of all,
we compare our API-Net with Baseline,
i.e.,
the standard ResNet-101 without our API design.
As expected,
API-Net largely outperforms it in Table \ref{tab:baseline},
showing the effectiveness of API module.

\textbf{Mutual Vector}.
To demonstrate the essentiality of $\mathbf{x}_{m}$ in Eq. (\ref{eq:MutualVector}),
we investigate different operations to generate it.
\textbf{(I)} \textit{Individual} Operation.
The key of $\mathbf{x}_{m}$ is to learn mutual information from both images in the pair.
For comparison,
we introduce a baseline without it.
Specifically,
we replace mutual learning in Eq. (\ref{eq:MutualVector}) by individual learning $\tilde{\mathbf{x}}_{i}=f_{m}(\mathbf{x}_{i})$,
and use $\tilde{\mathbf{x}}_{i}$ to generate the gate vector $\mathbf{g}_{i}=sigmoid(\tilde{\mathbf{x}}_{i})$ where $i\in\{1,2\}$.
\textbf{(II)} \textit{Compact Bilinear Pooling} Operation.
We generate $\mathbf{x}_{m}$ by compact bilinear pooling \cite{gao2016compact} between $\mathbf{x}_{1}$ and $\mathbf{x}_{2}$.
\textbf{(III)} \textit{Elementwise} Operations.
We perform a number of widely-used elementwise operations to generate $\mathbf{x}_{m}$,
including
\textit{Subtract Square} $\mathbf{x}_{m}=(\mathbf{x}_{1}-\mathbf{x}_{2})^{2}$,
\textit{Sum} $\mathbf{x}_{m}=\mathbf{x}_{1}+\mathbf{x}_{2}$,
and
\textit{Product} $\mathbf{x}_{m}=\mathbf{x}_{1}\odot\mathbf{x}_{2}$.
\textbf{(IV)} \textit{Weight Attention} Operation.
We use a two-layer MLP to generate the weight of $\mathbf{x}_{1}$ and $\mathbf{x}_{2}$,
i.e.,
$[w_{1}, w_{2}]=softmax(f_{w}([\mathbf{x}_{1}, \mathbf{x}_{2}]))$,
where
the output dimension of the 1st FC layer is 512.
Then,
we use the normalized weight vector as attention to generate the mutual vector,
i.e.,
$\mathbf{x}_{m}=\sum w_{i}\mathbf{x}_{i}$.
\textbf{(V)} \textit{MLP} Operation.
It is the mapping function described in the implementation details.
As shown in Table \ref{tab:NeceInter},
the \textit{Individual} operation (i.e., the setting without $\mathbf{x}_{m}$) performs worst.
Hence,
it is necessary to learn mutual context by $\mathbf{x}_{m}$,
which often plays an important role in finding distinct clues for individual image.
Moreover,
the performance of $\mathbf{x}_{m}$ operations is competitive.
We choose the simple but effective \textit{MLP} to generate the mutual vector in our experiments.

\begin{table}[t]
\begin{center}

\resizebox{0.6\columnwidth}{!}{
\begin{tabular}{l|cc}
\hline\hline
Method      &Baseline  & API-Net \\ \hline
Accuracy    &85.4  &\textbf{88.6}\\ \hline\hline
\end{tabular}
}
\end{center}
\caption{Comparison with basic methods.
Baseline is the standard ResNet-101 without our API design.
}
\label{tab:baseline}
%
\begin{center}
\resizebox{0.6\columnwidth}{!}{
\begin{tabular}{l|c}
\hline\hline
Mutual Vector             &Accuracy    \\ \hline
\textit{Individual}                        &87.9                 \\
\textit{Compact Bilinear Pooling}          &88.2                 \\
\textit{Subtract Square }                  &88.3                      \\
\textit{Sum}                               &88.5                     \\
\textit{Product }                          &88.4                     \\
\textit{Weight Attention }                 &88.4                  \\
\textit{MLP }                              &\textbf{88.6}         \\ \hline\hline
\end{tabular}
}
\end{center}
\caption{Different operations of mutual vector (CUB-200-2011).
More details can be found in Section \ref{Ablation Studies}.}
\label{tab:NeceInter}

\begin{center}
\resizebox{0.8\columnwidth}{!}{
\begin{tabular}{l|c||l|c}
\hline\hline
Gate Vector      &Accuracy        & Interaction & Accuracy    \\ \hline
\textit{Single}           & 87.7              & $\mathcal{L}_{ce}$         &  88.1     \\
\textit{Pair}             & \textbf{88.6}  & $\mathcal{L}_{ce}+\mathcal{L}_{rk}$     & \textbf{88.6} \\ \hline\hline
\end{tabular}
}

\end{center}
\caption{Different strategies of gate vector and interaction (CUB-200-2011).
More details can be found in Section \ref{Ablation Studies}.}
\label{tab:Interact}
\end{table}

\begin{table}[t]
\begin{center}
\resizebox{0.7\columnwidth}{!}{
\begin{tabular}{l|cc|c}
\hline\hline
Pair Construction & Intra & Inter & Accuracy \\ \hline
\textit{Random}                         & - & - & 86.4\\ \hline
\multirow{8}{*}{\textit{Class-Image}}   & - & D & 85.4\\
                                        & - & S & 87.2\\
                                        & D & - & 87.1 \\
                                        & S & - & 87.6 \\
                                        & D & D & 87.0\\
                                        & S & D & 87.4\\
                                        & D & S & 88.3\\
                                        & S & S & \textbf{88.6}\\ \hline\hline
\end{tabular}
}
\end{center}
\caption{Different strategies of pair construction (CUB-200-2011).
\textit{Random}:
We randomly sample 240 image pairs in a batch.
\textit{Class-Image}:
We sample 240 image pairs in a batch,
according to class (i.e., Intra / Inter) and Euclidean distance (i.e., Similar(S) / Dissimilar(D)).
This can generate 8 \textit{Class-Image} settings.
For example,
we randomly sample 30 classes in a batch.
For each class,
we randomly sample 4 images.
For each image,
we construct 2 pairs,
i.e.,
the intra / inter pair consists of this image and its most similar image from intra / inter classes.
This is denoted as Intra(S) \& Inter(S).
More explanations can be found in Section \ref{Ablation Studies}.
}
\label{tab:Sample}
\begin{center}
\resizebox{0.8\columnwidth}{!}{
\begin{tabular}{c|ccc}
\hline\hline
Class Size ($N_{cl}$)   & $N_{cl}$=10  & $N_{cl}$=20  & $N_{cl}$=30 \\ \hline
Accuracy     &   83.5  &   87.0  & \textbf{88.6}       \\ \hline\hline
Image Size ($N_{im}$)   & $N_{im}$=2   & $N_{im}$=3   & $N_{im}$=4 \\ \hline
Accuracy     & 88.1    &   88.2  & \textbf{88.6}       \\ \hline\hline
($N_{cl}$, $N_{im}$) & (24, 5) & (30, 4) & (40, 3) \\ \hline
Accuracy & 87.7 & \textbf{88.6} & 88.2 \\
\hline\hline
\end{tabular}
}
\end{center}
\caption{Class Size \& Image Size (CUB-200-2011).
After choosing Intra(S) \& Inter(S) in the \textit{Class-Image} rule,
we further explore the number of sampled classes and images in each batch.
When varying the class/image size,
we fix the image/class size as 4/30.}
\label{tab:Size}

\end{table}

\begin{table}[htb!]

\begin{center}
\resizebox{0.9\columnwidth}{!}{
\begin{tabular}{l|c|c|c}
\hline\hline
Method & Backbone & Extra S.  & CUB \\
\hline\hline
DeepLAC \cite{lin2015deep}& \multirow{2}{*}{AlexNet} & Yes & 80.3  \\
Part-RCNN \cite{zhang2014part}& & Yes & 81.6 \\
\hline

PA-CNN \cite{krause2015fine}& \multirow{2}{*}{VGGNet-19} & Yes & 82.8 \\
MG-CNN \cite{wang2015multiple}& & Yes & 83.0 \\
\hline

FCAN \cite{liu2016fully}& \multirow{3}{*}{ResNet-50} & Yes & 84.7 \\
TA-FGVC \cite{li2018read}& & Yes & 88.1 \\
HSE \cite{chen2018fine}& & Yes & 88.1 \\ \hline\hline
PDFR \cite{zhang2016picking} & \multirow{5}{*}{VGGNet-16} & No & 84.5 \\
Grassmann Pool\cite{wei2018grassmann} & & No & 85.8 \\
KP \cite{cui2017kernel} & & No & 86.2 \\
HBP \cite{yu2018hierarchical}& & No & 87.1 \\
$G^2$DeNet \cite{wang2017g2denet} & & No & 87.1 \\
\hline

MG-CNN \cite{wang2015multiple}&  \multirow{5}{*}{VGGNet-19} & No & 81.7 \\
B-CNN \cite{lin2015bilinear}& & No & 84.1 \\
RACNN \cite{fu2017look}& & No & 85.3 \\
MACNN \cite{zheng2017learning} & & No & 86.5 \\
Deep KSPD \cite{engin2018deepkspd} & & No & 86.5 \\
\hline

ST-CNN \cite{jaderberg2015spatial} & \multirow{1}{*}{GoogleNet} & No & 84.1 \\
\hline

FCAN \cite{liu2016fully}& \multirow{3}{*}{ResNet-50} & No & 84.3 \\
DFL-CNN \cite{wang2018learning}& & No & 87.4 \\
NTS-Net \cite{yang2018learning}& & No  & 87.5 \\
\hline

MAMC \cite{sun2018multi}& \multirow{3}{*}{ResNet-101} & No & 86.5 \\
TripletNet \cite{hoffer2015deep} & & No & 86.6 \\
iSQRT-COV \cite{li2018towards} & & No & 88.7 \\
\hline

MaxEnt \cite{dubey2018maximum} & \multirow{2}{*}{DenseNet-161} & No & 86.5 \\
PC \cite{dubey2018pairwise}&  & No & 86.9 \\
\hline
\textbf{Our API-Net} & \multirow{1}{*}{ResNet-50} & No & 87.7 \\
\textbf{Our API-Net} & \multirow{1}{*}{ResNet-101} & No & 88.6 \\
\textbf{Our API-Net} & \multirow{1}{*}{DenseNet-161} & No & \textbf{90.0} \\ \hline\hline
\end{tabular}
}
\end{center}
\caption{Comparison with The-State-of-The-Art (CUB-200-2011). Extra S.: Extra Supervision.}
\label{tab:sotaCUB}
%
%
\begin{center}
\resizebox{0.9\columnwidth}{!}{
\begin{tabular}{l|c|c|c}
\hline\hline
Method & Backbone & Extra S.  & Aircraft \\
\hline\hline
BoT \cite{wang2016mining}& \multirow{1}{*}{VGGNet-16} & Yes & 88.4\\

MG-CNN \cite{wang2015multiple}& \multirow{1}{*}{VGGNet-19} & Yes & 86.6 \\
 \hline\hline

KP \cite{cui2017kernel} & \multirow{6}{*}{VGGNet-16} & No & 86.9 \\
LRBP \cite{kong2017low} & & No & 87.3 \\
$G^2$DeNet \cite{wang2017g2denet} & & No & 89.0 \\
Grassmann Pool\cite{wei2018grassmann} & & No & 89.8 \\
HBP \cite{yu2018hierarchical}& & No  & 90.3 \\
DFL-CNN \cite{wang2018learning}& & No & 92.0 \\
\hline

B-CNN \cite{lin2015bilinear}& \multirow{4}{*}{VGGNet-19} & No & 84.1 \\
RACNN \cite{fu2017look}& & No & 88.4\\
MACNN \cite{zheng2017learning}& & No  & 89.9 \\
Deep KSPD \cite{engin2018deepkspd} & & No & 91.5 \\
\hline

NTS-Net \cite{yang2018learning}& \multirow{1}{*}{ResNet-50} & No  & 91.4 \\

iSQRT-COV \cite{li2018towards} & \multirow{1}{*}{ResNet-101} & No & 91.4 \\
\hline

PC \cite{dubey2018pairwise}& \multirow{2}{*}{DenseNet-161} & No & 89.2 \\
MaxEnt \cite{dubey2018maximum} &  & No & 89.8 \\
\hline

\hline
\textbf{Our API-Net} & \multirow{1}{*}{ResNet-50}  & No & 93.0 \\
\textbf{Our API-Net} & \multirow{1}{*}{ResNet-101}  & No & 93.4 \\
\textbf{Our API-Net} & \multirow{1}{*}{DenseNet-161}  & No &  \textbf{93.9}\\ \hline\hline
\end{tabular}
}
\end{center}
\caption{Comparison with The-State-of-The-Art (Aircraft). Extra S.: Extra Supervision.}
\label{tab:sotaAir}
\end{table}

\begin{table}[t]
\begin{center}
\resizebox{0.9\columnwidth}{!}{
\begin{tabular}{l|c|c|c}
\hline\hline
Method & Backbone & Extra S.  & Cars \\
\hline\hline
BoT \cite{wang2016mining}& \multirow{1}{*}{VGGNet-16} & Yes & 92.5\\

PA-CNN \cite{krause2015fine}& \multirow{1}{*}{VGGNet-19} & Yes & 92.8 \\
FCAN \cite{liu2016fully}& \multirow{1}{*}{ResNet-50} & Yes & 91.3 \\
\hline\hline
KP \cite{cui2017kernel} & \multirow{5}{*}{VGGNet-16} & No & 92.4 \\
$G^2$DeNet \cite{wang2017g2denet} & & No & 92.5 \\
Grassmann Pool\cite{wei2018grassmann} & & No & 92.8 \\
HBP \cite{yu2018hierarchical}& & No & 93.7 \\
DFL-CNN \cite{wang2018learning}& & No & 93.8 \\
\hline

B-CNN \cite{lin2015bilinear}& \multirow{4}{*}{VGGNet-19} & No & 91.3 \\
RACNN \cite{fu2017look}& & No & 92.5 \\
MACNN \cite{zheng2017learning}& & No & 92.8 \\
Deep KSPD \cite{engin2018deepkspd} & & No & 93.2 \\
\hline

FCAN \cite{liu2016fully}& \multirow{2}{*}{ResNet-50} & No & 89.1 \\
NTS-Net \cite{yang2018learning}& & No  & 93.9 \\
\hline

MAMC \cite{sun2018multi}& \multirow{2}{*}{ResNet-101} & No & 93.0 \\
iSQRT-COV \cite{li2018towards} &  & No & 93.3 \\
\hline


PC\cite{dubey2018pairwise}& \multirow{2}{*}{DenseNet-161} & No & 92.9 \\
MaxEnt \cite{dubey2018maximum} &  & No & 93.0 \\

 \hline
\textbf{Our API-Net}& ResNet-50 & No & 94.8 \\
\textbf{Our API-Net}& ResNet-101 & No & 94.9\\
\textbf{Our API-Net}& DenseNet-161 & No & \textbf{95.3} \\ \hline\hline
\end{tabular}
}
\end{center}
\caption{Comparison with The-State-of-The-Art (Stanford Cars). Extra S.: Extra Supervision.}
\label{tab:sotaCar}
%
%
%
\begin{center}
\resizebox{0.9\columnwidth}{!}{
\begin{tabular}{l|c|c|c}
\hline\hline
Method & Backbone & Extra S.  & Dogs \\
\hline\hline
TA-FGVC \cite{li2018read}& \multirow{1}{*}{ResNet-50} & Yes & 88.9 \\ \hline\hline
PDFR \cite{zhang2016picking} & \multirow{1}{*}{AlexNet} & No & 72.0 \\

DVAN \cite{zhao2017diversified} & \multirow{1}{*}{VGGNet-16} & No & 81.5 \\
\hline

B-CNN \cite{lin2015bilinear}& \multirow{2}{*}{VGGNet-19} & No & 82.1 \\
RACNN \cite{fu2017look}& & No & 87.3 \\
\hline

FCAN \cite{liu2016fully}& \multirow{1}{*}{ResNet-50} & No & 84.2 \\


MAMC \cite{sun2018multi}& \multirow{1}{*}{ResNet-101} & No & 85.2 \\
\hline

MaxEnt \cite{dubey2018maximum} &  \multirow{2}{*}{DenseNet-161} & No & 83.6 \\
PC \cite{dubey2018pairwise}&  & No & 83.8\\
 \hline
\textbf{Our API-Net}& \multirow{1}{*}{ResNet-50} & No & 88.3\\
\textbf{Our API-Net}& \multirow{1}{*}{ResNet-101} & No & \textbf{90.3}\\
\textbf{Our API-Net}& \multirow{1}{*}{DenseNet-161} & No & 89.4\\ \hline\hline
\end{tabular}
}
\end{center}
\caption{Comparison with The-State-of-The-Art (Stanford Dogs). Extra S.: Extra Supervision.}
\label{tab:sotaDogs}

\begin{center}
\resizebox{0.9\columnwidth}{!}{
\begin{tabular}{l|c|c|c}
  \hline \hline
  Method & Backbone & Extra S.  & NABirds \\
  \hline \hline
  Van et al. \cite{van2015building} & AlexNet & Yes & 75.0 \\
  \hline \hline
  Branson et al. \cite{branson2014bird} & AlexNet & No & 35.7 \\
  \hline
  PC \cite{dubey2018pairwise}&  \multirow{2}{*}{DenseNet-161} & No & 82.8\\
  MaxEnt \cite{dubey2018maximum} &  & No & 83.0 \\
  \hline
  \textbf{Our API-Net}& \multirow{1}{*}{ResNet-50} & No & 86.2\\
  \textbf{Our API-Net}& \multirow{1}{*}{ResNet-101} & No & 86.6\\
  \textbf{Our API-Net}& \multirow{1}{*}{DenseNet-161} & No & \textbf{88.1}\\ \hline\hline
\end{tabular}
}
\end{center}
\caption{Comparison with The-State-of-The-Art (NABirds). Extra S.: Extra Supervision.}
\label{tab:sotaNABirds}

\end{table}

\textbf{Gate Vector}.
We discuss different approaches to generate the gate vector.
\textbf{(I)} \textit{Single}.
Since $\mathbf{x}_{m}$ inherits mutual characteristics from both $\mathbf{x}_{1}$ and $\mathbf{x}_{2}$,
a straightforward choice is to use $\mathbf{g}_{m}=sigmoid(\mathbf{x}_{m})$ as a single gate vector.
Subsequently,
one can train API-Net with attentive features $\mathbf{x}_{i}^{self}=\mathbf{x}_{i}+\mathbf{x}_{i}\odot\mathbf{g}_{m}$,
where $i\in\{1,2\}$.
\textbf{(II)} \textit{Pair}.
This is the proposed setting in Eq. (\ref{eq:GateVector}).
As shown in Table \ref{tab:Interact},
the \textit{Pair} setting outperforms the \textit{Single} setting.
It illustrates that,
we have to discover discriminative clues from the view of each image,
by comparing $\mathbf{x}_{i}$ with $\mathbf{x}_{m}$.

\textbf{Interaction}.
We explore different interaction strategies.
\textbf{(I)} $\mathcal{L}_{ce}$. 
We train API-Net only with cross entropy loss $\mathcal{L}_{ce}$.
In this case,
score-ranking regularization $\mathcal{L}_{rk}$ is not used for training.
\textbf{(II)} $\mathcal{L}_{ce}+\mathcal{L}_{rk}$. 
We train API-Net with the proposed loss $\mathcal{L}$.
In Table \ref{tab:Interact},
our proposed loss performs better.
It illustrates that,
$\mathcal{L}_{rk}$ can further generalize pairwise interaction with different feature priorities.

\textbf{Pair Construction}.
We investigate different strategies to construct input image pairs.
\textbf{(I)} \textit{Random}.
We randomly sample 240 image pairs in a batch.
\textbf{(II)} \textit{Class-Image}.
We sample 240 image pairs,
according to class (i.e., Intra / Inter) and Euclidean distance between features (i.e., Similar(S) / Dissimilar(D)).
This can generate 8 \textit{Class-Image} settings in Table \ref{tab:Sample}.
For example,
we randomly sample 30 classes in a batch.
For each class,
we randomly sample 4 images.
For each image,
we construct 2 pairs,
i.e.,
the intra / inter pair consists of this image and its most similar image from intra / inter classes.
This is denoted as Intra(S) \& Inter(S).
The results of different settings are shown in Table \ref{tab:Sample}.
First,
most \textit{Class-Image} settings outperform the \textit{Random} setting.
It illustrates that,
one should take the prior knowledge of class and similarity into account,
when constructing image pairs.
Second,
Inter(S) outperforms Inter(D),
no matter which the intra setting is.
The reason is that,
each pair in Inter(S) / Inter(D) consists of two most similar / dissimilar images from different classes,
i.e.,
the pairs in Inter(S) increases the difficulty of both being recognized correctly at the same time.
By checking such pairs in Inter(S),
API-Net can be trained to distinguish subtle semantic differences.
Third,
the performance is competitive between Intra(S) and Intra(D),
no matter which the inter setting is.
This is mainly because intra pairs are with the same label.
API-Net does not need to put much effort to recognize why these pairs are similar.
Finally,
the settings with both intra and inter pairs outperform those with only intra or inter pairs.
It is credited to the fact that,
API-Net can gradually distinguish which are highly-confused or truly-similar pairs,
by leveraging both intra and inter pairs.
We choose the setting with the best performance,
i.e.,
Intra(S) \& Inter(S) in our experiment.

\textbf{Class Size \& Image Size}.
After choosing Intra(S) \& Inter(S) in the \textit{Class-Image} setting,
we further explore the number of sampled classes and images in each batch.
When varying the class/image size,
we fix the image/class size as 4/30.
The results are shown in Table \ref{tab:Size}.
One can see that,
API-Net is more sensitive to class size than image size.
The main reason is that,
more classes often produce richer diversity of images pairs.
We choose the best setting in our experiment,
i.e.,
class size=30 and image size=4.

\begin{figure*}[htb!]
\begin{center}

\includegraphics[width=1.96\columnwidth, height=1.20\columnwidth]{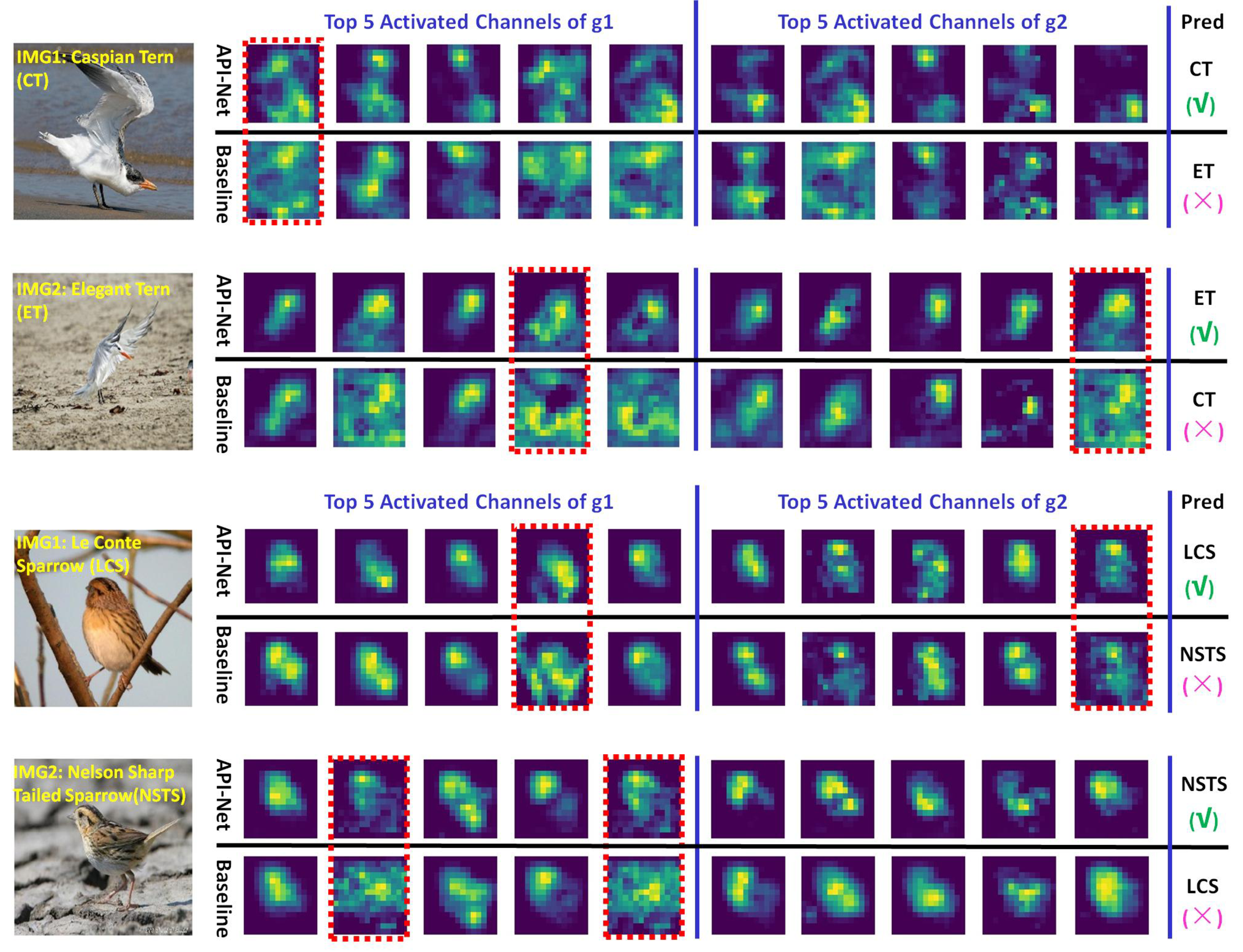}
\end{center}
\caption{Visualization.
As shown in red-dashed boxes,
many feature maps of Baseline are confused or noisy,
e.g.,
the object regions are blurring,
or certain background regions are activated.
On the contrary,
our API-Net can effectively discover and then distinguish discriminative clues via attentive pairwise interaction.
More explanations can be found in Section \ref{Visualization}.
}
\label{fig:visual}
\end{figure*}

\subsection{Comparison with The-State-of-The-Art}

We compare API-Net with a number of recent works on five widely-used benchmarks in Table \ref{tab:sotaCUB}-\ref{tab:sotaNABirds}.
First,
API-Net outperforms the object-part-localization approaches such as RACNN \cite{fu2017look} and MACNN \cite{zheng2017learning},
showing the importance of API-Net for discriminative feature learning.
Second,
API-Net outperforms the patch-interaction approaches such as B-CNN \cite{lin2015bilinear},
HBP \cite{yu2018hierarchical}.
It demonstrates that,
we need to put more efforts to distinguish two different images jointly,
instead of learning with individual image alone.
Third,
API-Net outperforms the metric-learning approaches such as PC \cite{dubey2018pairwise}.
Particularly, we also re-implement TripletNet \cite{hoffer2015deep} in Table \ref{tab:sotaCUB}, a classical approach optimized by extra triplet loss,
and compare it with API-Net.
They all illustrate that,
we should further exploit how to discover differential visual clues in the image pair,
instead of straightforwardly regularizing the sample distribution in the feature space.
Fourth,
API-Net even outperforms the approaches with extra supervision such as
localization annotations \cite{zhang2014part},
super-category labels \cite{chen2018fine},
or text \cite{li2018read}.
It shows the effectiveness of API module.
Finally,
without complex architectures and multi-stage learning in the previous works,
our API module can be easily integrated into the standard CNN training procedure
and serve as a plug-and-play unit for discovering subtle visual differences.
Hence,
API-Net is a concise and practical deep learning approach for fine-grained classification.

\subsection{Visualization}
\label{Visualization}

We further visualize API-Net in Fig. \ref{fig:visual}.
First,
we choose two pairs from the highly-confused categories in CUB-200-2011,
i.e.,
\textit{Caspian Tern}~(CT) vs. \textit{Elegant Tern}~(ET),
and
\textit{Le Conte Sparrow}~(LCS) vs. \textit{Nelson Sharp Tailed Sparrow}~(NSTS).
Second,
we feed each pair into API-Net,
and find top-5 activated channels of gate vectors $\mathbf{g}_{1}$ and $\mathbf{g}_{2}$.
Subsequently,
we show the corresponding feature maps ($14\times14$) before global pooling.
Additionally,
we show the corresponding feature maps in Baseline,
i.e.,
ResNet-101 without attentive pairwise interaction.

As shown in red-dashed boxes of Fig. \ref{fig:visual},
many feature maps of Baseline are confused or noisy,
e.g.,
the object regions are blurring,
or certain background regions are activated.
On the contrary,
our API-Net can effectively discover distinct features,
e.g.,
$\mathbf{g}_{1}$ mainly focuses on the body of \textit{Caspian Tern},
while
$\mathbf{g}_{2}$ mainly focuses on the mouth of \textit{Elegant Tern}.
These contrastive clues allow API to correctly distinguish such two birds.

Additionally,
it is interesting to mention that,
API-Net can automatically pay attention to discriminative object parts in the feature maps,
even though it mainly works on high-level feature vectors. 
This observation indicates that,
CNN can be well generalized via attentive pairwise interaction.
As expected,
our API-Net successfully recognizes all these pairs,
while Baseline makes wrong predictions.

\section{Conclusion}
In this paper,
we propose a novel Attentive Pairwise Interaction Network (API-Net) for fine-grained classification.
It can adaptively discover contrastive cues from a pair of images,
and attentively distinguish them via pairwise interaction.
The results on five popular fine-grained benchmarks show that,
API-Net achieves the state-of-the-art performance.

\section{ Acknowledgments}
This work is partially supported by the National Key Research and Development Program of China (No. 2016YFC1400704), and National Natural Science Foundation of China (61876176, U1713208), Shenzhen Basic Research Program (JCYJ20170818164704758, CXB201104220032A), the Joint Lab of CAS-HK, Shenzhen Institute of Artificial Intelligence and Robotics for Society.


{\small
\bibliographystyle{aaai}
\bibliography{2505_Bibliography_File}

\begin{thebibliography}{}

\bibitem[\protect\citeauthoryear{Branson \bgroup et al\mbox.\egroup
  }{2014}]{branson2014bird}
Branson, S.; Van~Horn, G.; Belongie, S.; and Perona, P.
\newblock 2014.
\newblock Bird species categorization using pose normalized deep convolutional
  nets.
\newblock {\em arXiv preprint arXiv:1406.2952}.

\bibitem[\protect\citeauthoryear{Bruner}{2017}]{bruner2017study}
Bruner, J.
\newblock 2017.
\newblock {\em A study of thinking}.
\newblock Routledge.

\bibitem[\protect\citeauthoryear{Cai, Zuo, and Zhang}{2017}]{cai2017higher}
Cai, S.; Zuo, W.; and Zhang, L.
\newblock 2017.
\newblock Higher-order integration of hierarchical convolutional activations
  for fine-grained visual categorization.
\newblock In {\em ICCV},  511--520.

\bibitem[\protect\citeauthoryear{Chen \bgroup et al\mbox.\egroup
  }{2018}]{chen2018fine}
Chen, T.; Wu, W.; Gao, Y.; Dong, L.; Luo, X.; and Lin, L.
\newblock 2018.
\newblock Fine-grained representation learning and recognition by exploiting
  hierarchical semantic embedding.
\newblock {\em arXiv preprint arXiv:1808.04505}.

\bibitem[\protect\citeauthoryear{Cui \bgroup et al\mbox.\egroup
  }{2017}]{cui2017kernel}
Cui, Y.; Zhou, F.; Wang, J.; Liu, X.; Lin, Y.; and Belongie, S.
\newblock 2017.
\newblock Kernel pooling for convolutional neural networks.
\newblock In {\em CVPR},  2921--2930.

\bibitem[\protect\citeauthoryear{Dubey \bgroup et al\mbox.\egroup
  }{2018a}]{dubey2018pairwise}
Dubey, A.; Gupta, O.; Guo, P.; Raskar, R.; Farrell, R.; and Naik, N.
\newblock 2018a.
\newblock Pairwise confusion for fine-grained visual classification.
\newblock In {\em ECCV},  70--86.

\bibitem[\protect\citeauthoryear{Dubey \bgroup et al\mbox.\egroup
  }{2018b}]{dubey2018maximum}
Dubey, A.; Gupta, O.; Raskar, R.; and Naik, N.
\newblock 2018b.
\newblock Maximum-entropy fine grained classification.
\newblock In {\em NIPS}.

\bibitem[\protect\citeauthoryear{Engin \bgroup et al\mbox.\egroup
  }{2018}]{engin2018deepkspd}
Engin, M.; Wang, L.; Zhou, L.; and Liu, X.
\newblock 2018.
\newblock Deepkspd: learning kernel-matrix-based spd representation for
  fine-grained image recognition.
\newblock In {\em ECCV},  612--627.

\bibitem[\protect\citeauthoryear{Fu, Zheng, and Mei}{2017}]{fu2017look}
Fu, J.; Zheng, H.; and Mei, T.
\newblock 2017.
\newblock Look closer to see better: Recurrent attention convolutional neural
  network for fine-grained image recognition.
\newblock In {\em CVPR},  4438--4446.

\bibitem[\protect\citeauthoryear{Gao \bgroup et al\mbox.\egroup
  }{2016}]{gao2016compact}
Gao, Y.; Beijbom, O.; Zhang, N.; and Darrell, T.
\newblock 2016.
\newblock Compact bilinear pooling.
\newblock In {\em CVPR},  317--326.

\bibitem[\protect\citeauthoryear{He \bgroup et al\mbox.\egroup
  }{2016}]{he2016deep}
He, K.; Zhang, X.; Ren, S.; and Sun, J.
\newblock 2016.
\newblock Deep residual learning for image recognition.
\newblock In {\em CVPR},  770--778.

\bibitem[\protect\citeauthoryear{Hermans, Beyer, and
  Leibe}{2017}]{hermans2017defense}
Hermans, A.; Beyer, L.; and Leibe, B.
\newblock 2017.
\newblock In defense of the triplet loss for person re-identification.
\newblock {\em arXiv preprint arXiv:1703.07737}.

\bibitem[\protect\citeauthoryear{Hoffer and Ailon}{2015}]{hoffer2015deep}
Hoffer, E., and Ailon, N.
\newblock 2015.
\newblock Deep metric learning using triplet network.
\newblock In {\em International Workshop on Similarity-Based Pattern
  Recognition},  84--92.
\newblock Springer.

\bibitem[\protect\citeauthoryear{Huang \bgroup et al\mbox.\egroup
  }{2017}]{huang2017densely}
Huang, G.; Liu, Z.; Van Der~Maaten, L.; and Weinberger, K.~Q.
\newblock 2017.
\newblock Densely connected convolutional networks.
\newblock In {\em CVPR},  4700--4708.

\bibitem[\protect\citeauthoryear{Jaderberg \bgroup et al\mbox.\egroup
  }{2015}]{jaderberg2015spatial}
Jaderberg, M.; Simonyan, K.; Zisserman, A.; et~al.
\newblock 2015.
\newblock Spatial transformer networks.
\newblock In {\em NIPS},  2017--2025.

\bibitem[\protect\citeauthoryear{Khosla \bgroup et al\mbox.\egroup
  }{2011}]{khosla2011novel}
Khosla, A.; Jayadevaprakash, N.; Yao, B.; and Li, F.-F.
\newblock 2011.
\newblock Novel dataset for fine-grained image categorization: Stanford dogs.
\newblock In {\em CVPR Workshops}.

\bibitem[\protect\citeauthoryear{Kong and Fowlkes}{2017}]{kong2017low}
Kong, S., and Fowlkes, C.
\newblock 2017.
\newblock Low-rank bilinear pooling for fine-grained classification.
\newblock In {\em CVPR},  365--374.

\bibitem[\protect\citeauthoryear{Krause \bgroup et al\mbox.\egroup
  }{2013}]{krause20133d}
Krause, J.; Stark, M.; Deng, J.; and Fei-Fei, L.
\newblock 2013.
\newblock 3d object representations for fine-grained categorization.
\newblock In {\em ICCV Workshops},  554--561.

\bibitem[\protect\citeauthoryear{Krause \bgroup et al\mbox.\egroup
  }{2015}]{krause2015fine}
Krause, J.; Jin, H.; Yang, J.; and Fei-Fei, L.
\newblock 2015.
\newblock Fine-grained recognition without part annotations.
\newblock In {\em CVPR},  5546--5555.

\bibitem[\protect\citeauthoryear{Kulis and others}{2013}]{kulis2013metric}
Kulis, B., et~al.
\newblock 2013.
\newblock Metric learning: A survey.
\newblock {\em Foundations and Trends{\textregistered} in Machine Learning}
  5(4):287--364.

\bibitem[\protect\citeauthoryear{Li \bgroup et al\mbox.\egroup
  }{2018a}]{li2018read}
Li, J.; Zhu, L.; Huang, Z.; Lu, K.; and Zhao, J.
\newblock 2018a.
\newblock I read, i saw, i tell: texts assisted fine-grained visual
  classification.
\newblock In {\em ACMMM},  663--671.

\bibitem[\protect\citeauthoryear{Li \bgroup et al\mbox.\egroup
  }{2018b}]{li2018towards}
Li, P.; Xie, J.; Wang, Q.; and Gao, Z.
\newblock 2018b.
\newblock Towards faster training of global covariance pooling networks by
  iterative matrix square root normalization.
\newblock In {\em CVPR},  947--955.

\bibitem[\protect\citeauthoryear{Lin \bgroup et al\mbox.\egroup
  }{2015}]{lin2015deep}
Lin, D.; Shen, X.; Lu, C.; and Jia, J.
\newblock 2015.
\newblock Deep lac: Deep localization, alignment and classification for
  fine-grained recognition.
\newblock In {\em CVPR},  1666--1674.

\bibitem[\protect\citeauthoryear{Lin, RoyChowdhury, and
  Maji}{2015}]{lin2015bilinear}
Lin, T.-Y.; RoyChowdhury, A.; and Maji, S.
\newblock 2015.
\newblock Bilinear cnn models for fine-grained visual recognition.
\newblock In {\em ICCV},  1449--1457.

\bibitem[\protect\citeauthoryear{Liu \bgroup et al\mbox.\egroup
  }{2016}]{liu2016fully}
Liu, X.; Xia, T.; Wang, J.; Yang, Y.; Zhou, F.; and Lin, Y.
\newblock 2016.
\newblock Fully convolutional attention networks for fine-grained recognition.
\newblock {\em arXiv preprint arXiv:1603.06765}.

\bibitem[\protect\citeauthoryear{Maji \bgroup et al\mbox.\egroup
  }{2013}]{maji2013fine}
Maji, S.; Rahtu, E.; Kannala, J.; Blaschko, M.; and Vedaldi, A.
\newblock 2013.
\newblock Fine-grained visual classification of aircraft.
\newblock {\em arXiv preprint arXiv:1306.5151}.

\bibitem[\protect\citeauthoryear{Schroff, Kalenichenko, and
  Philbin}{2015}]{schroff2015facenet}
Schroff, F.; Kalenichenko, D.; and Philbin, J.
\newblock 2015.
\newblock Facenet: A unified embedding for face recognition and clustering.
\newblock In {\em CVPR},  815--823.

\bibitem[\protect\citeauthoryear{Sun \bgroup et al\mbox.\egroup
  }{2018}]{sun2018multi}
Sun, M.; Yuan, Y.; Zhou, F.; and Ding, E.
\newblock 2018.
\newblock Multi-attention multi-class constraint for fine-grained image
  recognition.
\newblock In {\em ECCV},  805--821.

\bibitem[\protect\citeauthoryear{Van~Horn \bgroup et al\mbox.\egroup
  }{2015}]{van2015building}
Van~Horn, G.; Branson, S.; Farrell, R.; Haber, S.; Barry, J.; Ipeirotis, P.;
  Perona, P.; and Belongie, S.
\newblock 2015.
\newblock Building a bird recognition app and large scale dataset with citizen
  scientists: The fine print in fine-grained dataset collection.
\newblock In {\em CVPR},  595--604.

\bibitem[\protect\citeauthoryear{Wah \bgroup et al\mbox.\egroup
  }{2011}]{wah2011caltech}
Wah, C.; Branson, S.; Welinder, P.; Perona, P.; and Belongie, S.
\newblock 2011.
\newblock The caltech-ucsd birds-200-2011 dataset.
\newblock {\em California Institute of Technology}.

\bibitem[\protect\citeauthoryear{Wang \bgroup et al\mbox.\egroup
  }{2015}]{wang2015multiple}
Wang, D.; Shen, Z.; Shao, J.; Zhang, W.; Xue, X.; and Zhang, Z.
\newblock 2015.
\newblock Multiple granularity descriptors for fine-grained categorization.
\newblock In {\em ICCV},  2399--2406.

\bibitem[\protect\citeauthoryear{Wang \bgroup et al\mbox.\egroup
  }{2016}]{wang2016mining}
Wang, Y.; Choi, J.; Morariu, V.; and Davis, L.~S.
\newblock 2016.
\newblock Mining discriminative triplets of patches for fine-grained
  classification.
\newblock In {\em CVPR},  1163--1172.

\bibitem[\protect\citeauthoryear{Wang, Li, and Zhang}{2017}]{wang2017g2denet}
Wang, Q.; Li, P.; and Zhang, L.
\newblock 2017.
\newblock G2denet: Global gaussian distribution embedding network and its
  application to visual recognition.
\newblock In {\em CVPR},  2730--2739.

\bibitem[\protect\citeauthoryear{Wang, Morariu, and
  Davis}{2018}]{wang2018learning}
Wang, Y.; Morariu, V.~I.; and Davis, L.~S.
\newblock 2018.
\newblock Learning a discriminative filter bank within a cnn for fine-grained
  recognition.
\newblock In {\em CVPR},  4148--4157.

\bibitem[\protect\citeauthoryear{Wei \bgroup et al\mbox.\egroup
  }{2018}]{wei2018grassmann}
Wei, X.; Zhang, Y.; Gong, Y.; Zhang, J.; and Zheng, N.
\newblock 2018.
\newblock Grassmann pooling as compact homogeneous bilinear pooling for
  fine-grained visual classification.
\newblock In {\em ECCV},  355--370.

\bibitem[\protect\citeauthoryear{Yang \bgroup et al\mbox.\egroup
  }{2018}]{yang2018learning}
Yang, Z.; Luo, T.; Wang, D.; Hu, Z.; Gao, J.; and Wang, L.
\newblock 2018.
\newblock Learning to navigate for fine-grained classification.
\newblock In {\em ECCV},  420--435.

\bibitem[\protect\citeauthoryear{Yu \bgroup et al\mbox.\egroup
  }{2018}]{yu2018hierarchical}
Yu, C.; Zhao, X.; Zheng, Q.; Zhang, P.; and You, X.
\newblock 2018.
\newblock Hierarchical bilinear pooling for fine-grained visual recognition.
\newblock In {\em ECCV},  574--589.

\bibitem[\protect\citeauthoryear{Zhang \bgroup et al\mbox.\egroup
  }{2014}]{zhang2014part}
Zhang, N.; Donahue, J.; Girshick, R.; and Darrell, T.
\newblock 2014.
\newblock Part-based r-cnns for fine-grained category detection.
\newblock In {\em ECCV},  834--849.
\newblock Springer.

\bibitem[\protect\citeauthoryear{Zhang \bgroup et al\mbox.\egroup
  }{2016a}]{zhang2016embedding}
Zhang, X.; Zhou, F.; Lin, Y.; and Zhang, S.
\newblock 2016a.
\newblock Embedding label structures for fine-grained feature representation.
\newblock In {\em CVPR},  1114--1123.

\bibitem[\protect\citeauthoryear{Zhang \bgroup et al\mbox.\egroup
  }{2016b}]{zhang2016picking}
Zhang, X.; Xiong, H.; Zhou, W.; Lin, W.; and Tian, Q.
\newblock 2016b.
\newblock Picking deep filter responses for fine-grained image recognition.
\newblock In {\em CVPR},  1134--1142.

\bibitem[\protect\citeauthoryear{Zhao \bgroup et al\mbox.\egroup
  }{2017}]{zhao2017diversified}
Zhao, B.; Wu, X.; Feng, J.; Peng, Q.; and Yan, S.
\newblock 2017.
\newblock Diversified visual attention networks for fine-grained object
  classification.
\newblock {\em TMM} 19(6):1245--1256.

\bibitem[\protect\citeauthoryear{Zheng \bgroup et al\mbox.\egroup
  }{2017}]{zheng2017learning}
Zheng, H.; Fu, J.; Mei, T.; and Luo, J.
\newblock 2017.
\newblock Learning multi-attention convolutional neural network for
  fine-grained image recognition.
\newblock In {\em ICCV},  5209--5217.

\end{thebibliography}
}

\end{document}